\documentclass[conference]{IEEEtran}
\IEEEoverridecommandlockouts
\usepackage{cite}
\usepackage{amsmath,amssymb,amsfonts}
\usepackage{algorithmic}
\usepackage{graphicx}
\usepackage{textcomp}
\usepackage{xcolor}
\usepackage{tabularx}
\usepackage{multirow}
\usepackage{algorithm, algorithmic}

\def\BibTeX{{\rm B\kern-.05em{\sc i\kern-.025em b}\kern-.08em
    T\kern-.1667em\lower.7ex\hbox{E}\kern-.125emX}}

\begin{document}
\title{Temporal Tensor Transformation Network for Multivariate Time Series Prediction}

\author{
    \IEEEauthorblockN{
        Yuya Jeremy Ong\IEEEauthorrefmark{1},
        Mu Qiao\IEEEauthorrefmark{2}, 
        Divyesh Jadav\IEEEauthorrefmark{3}}
    \IEEEauthorblockA{
        IBM Almaden Research Center, San Jose, CA, USA\\
        Email: 
            \IEEEauthorrefmark{1}yuyajong@ibm.com,
            \{\IEEEauthorrefmark{2}mqiao, \IEEEauthorrefmark{3}divyesh\}@us.ibm.com
    }
}

\maketitle

\begin{abstract}
Multivariate time series prediction has applications in a wide variety of domains and is considered to be a very challenging task, especially when the variables have correlations and exhibit complex temporal patterns, such as seasonality and trend. Many existing methods suffer from strong statistical assumptions, numerical issues with high dimensionality, manual feature engineering efforts, and scalability. In this work, we present a novel deep learning architecture, known as Temporal Tensor Transformation Network, which transforms the original multivariate time series into a higher order of tensor through the proposed Temporal-Slicing Stack Transformation. This yields a new representation of the original multivariate time series, which enables the convolution kernel to extract complex and non-linear features as well as variable interactional signals from a relatively large temporal region. Experimental results show that Temporal Tensor Transformation Network outperforms several state-of-the-art methods on window-based predictions across various tasks. The proposed architecture also demonstrates robust prediction performance through an extensive sensitivity analysis.

\end{abstract}


\section{Introduction} 
Multivariate time series analysis has gained wide spread applications in many fields, e.g., financial market prediction, weather forecasting, and energy consumption prediction. It is used to model and explain the underlying temporal patterns among a group of time series variables in dynamical systems. Various methods have been proposed to predict multivariate time series based on statistical modeling and deep neural networks.

Classical statistical models assume that the time series is stationary, i.e., the summary statistics of data points are consistent over time. Preprocessing procedures are usually needed to remove trend, seasonality, and other time-dependent structures from the raw series in order to make the data stationary. In addition, these models also assume the independence condition in the underlying linear regression problem, i.e., the random errors in the model are not correlated over time. Autocorrelation and partial autocorrelation functions are usually applied to identify the appropriate order of variables. Constructing statistically meaningful prediction models requires performing various preprocessing, transformations, and feature engineering, which are time consuming and difficult to scale. On the other hand, deep learning-based approaches, such as recurrent neural network, have demonstrated state-of-the-art performance in modeling time series data  through the use of stateful models. Recently, convolutional neural network (CNN) has also been applied to predict multivariate time series. Specifically, CNN is used as a stateless model to directly extract features from raw time series and generate predictions, or is used as part of the feature extraction step within a RNN architecture. 

Existing work of using CNN for multivariate time series prediction treats the time series as an image. For example, the number of variables\footnote{The terms ``variable'' and ``feature'' are used interchangeably in this paper.} is equal to the width of the image while the number of time steps is equal to the length of the image. The convolution is conducted over the temporal-variable plane. In this context, one of the key underlying structural premises of the convolution operation assumes a local spatially dependent topology of the data, as opposed to the dense layer where all the inputs are jointly modeled. That is, in a convolutional layer, neurons receive input from a restricted sub-area of the previous layer (aka the receptive field), whereas neurons in the dense layer receives input from the entire previous layer. Therefore, when the CNN kernel convolves over the variable-temporal plane, it is only able to observe a narrow set of variable interactions within a given time window. Alternatively, convolution operations where we consider each feature in the time series as an independent channel also suffers from similar issues where the kernel is only able to  observe a small local-window of time steps as its filters convolve over the data. In both scenarios, the limited view of the receptive field renders a local focus of the time series. Different from image processing, where objects in different regions could be quite distinct, time series data tend to be relatively ``homogeneous''. The prediction of future values depend more on the global pattern within the historical time window, rather than a local pattern. In this paper, we propose a novel neural network architecture for multivariate time series prediction with a new class of transformation functions known as \textit{temporal-slicing stack transformation}. These operations transform the original input data structure into a higher-order tensor, where the individual features in the time series are rearranged from a 1D temporal sequence into a 2D matrix. This transformation expands the view of the receptive field. As a result, the convolution is operated on a temporally larger region, which may help capture time-dependent features such as trend and seasonality, as well as variable interactions in a longer range.

\begin{figure}[ht]
\centering
\includegraphics[width=0.5\textwidth]{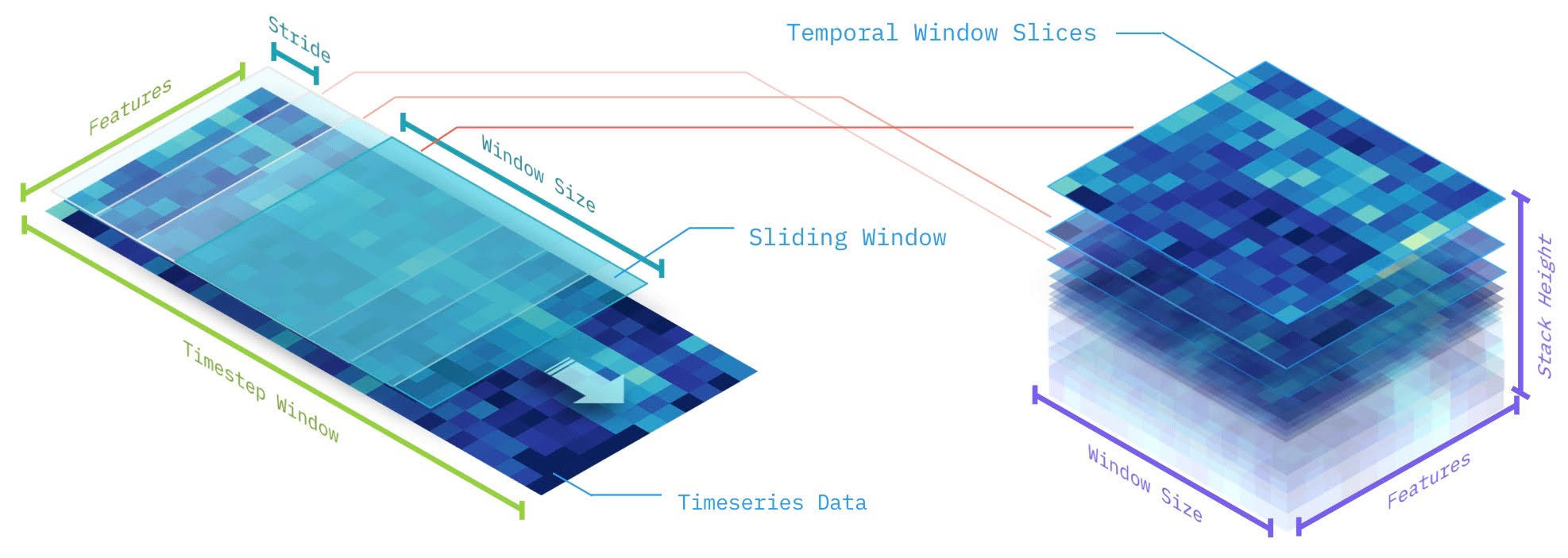}
\centering{\caption{Visualization of the Temporal-Slicing Stack Transformation process}\label{tt_visual}}
\end{figure}

\begin{figure*}[ht!]
\centering
\includegraphics[width=\textwidth]{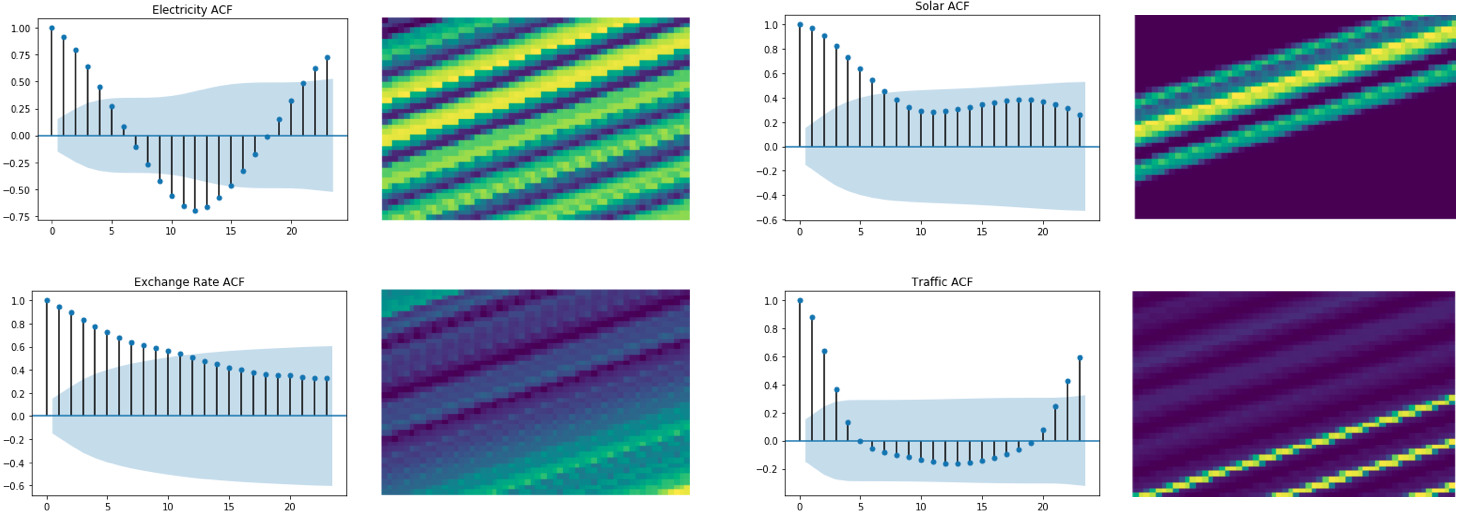}
\centering{\caption{Comparison of the autocorrelation plots and the resulting Temporal Tensor Transformations}\label{tt_acf}}
\end{figure*}

The temporal-slicing stack transformation is demonstrated by Figure \ref{tt_visual}, where a window is utilized to slide over the 2D time series data and extract a collection of time series slices. These slices are then correspondingly stacked on top of each other sequentially to form a 3D tensor. Specifically the three dimensions are ``features'', ``time'', and ``stack''. For multivariate time series, the transformation will convert a 2D time series into a 3D tensor. For univariate time series, the transformation will render a 2D matrix. The resulting structure yields an emergent pattern, where a spatial feature extractor,  such as CNN, can explicitly model complex and nonlinear autocorrelational-like features. To illustrate this, we use the transformation of univariate time series as an example. Figure \ref{tt_acf} shows the transformation results of univariate time series from four different datasets. We also show their corresponding autocorrelation plots. The transformation demonstrates strong patterns for time series of high autocorrelations. Comparing with the original 1D time series, the 2D matrix has much richer information and therefore may enable more informative feature extraction. 

The following sections of this paper are organized as follows. We highlight various statistical and deep learning approaches for time series analysis in Section II. Definitions of the temporal tensor transformation, the temporal-slicing stack transformation, and the proposed model architecture are described in Section III. The experimental setup, dataset, metrics, and results are presented in Section IV. In Section V, we perform sensitivity analysis based on experiments with synthesized datasets and controlling the input and output dimensions. We finally conclude and discuss future directions in Section VI.
\section{Related Work}
In this section, we review the prior work for multivariate time series prediction. We first introduce some of the known classical statistical methods and then present some of the state-of-the-art deep learning methods for time series modeling.

Time series prediction has been well studied with various statistical modeling methods for both univariate and multivariate data. For univariate time series modeling, the Auto Regressive Integrated Moving Average (ARIMA) model is often applied using the Box-Jenkins modeling process \cite{box2015time}. Other variants of the ARIMA models have been proposed to model temporal patterns of the data, for example, seasonality (SARMA) and coefficient dependent periodicity (PARMA) \cite{hannan1955test}. Furthermore, methods combining ARIMA and neural networks have also been developed to model both linear and non-linear dynamics \cite{zhang2003time}. For multivariate time series prediction, vector based methods, such as Vector Auto Regression (VAR) methods, extend the Auto Regressive (AR) models for univariate time series modeling \cite{box2015time} \cite{lutkepohl2005new}. Variants of the VAR model have also been proposed, such as the VARMAX model  \cite{milhoj2016multiple} and the VARX model \cite{bierens2004var}, where the model subsumes properties of the original VAR model and jointly learns interactions between the given variables.  Additionally, Gaussian Process \cite{alvarez2009sparse}, as a non-parametric statistical model, is used to predict over a distribution of continuous variables, as opposed to the aforementioned parametric methods. Statistical models often have very high computational complexity and face numerical issues, when the number of variables in the time series is high. 

With the advent of deep learning, new neural network architectures have been proposed for multivariate time series prediction. Borovykh et al. \cite{borovykh2017conditional} develop a multivariate time series model based on the WaveNet architecture \cite{van2016wavenet}, which is originally designed for speech audio signal processing. They augment the original architecture by simplifying and optimizing its core algorithms with the dialated convolution to capture long-term multivariate temporal data with noisy signals. Another multivariate time series modeling framework is proposed by Lai et al. \cite{lai2018modeling}, namely LSTNet, which combines CNN and RNN to extract hierarchical short-term and long-term temporal dependencies from the time series. In addition to model dependencies, LSTNet also accounts for the autoregressive component of the model as a residual connection between the CNN and LSTM components. Qin et al. \cite{qin2017dual} develop a dual-stage attention-based neural network architecture for multivariate time series modeling, which utilizes two sets of RNN as an encoder-decoder based architecture. Through each stage of the attention-based network architecture, the method attends to both the feature and temporal dimensions to adaptively select the relevant driving series.
\section{Model Architecture}
In this section, we present the proposed \textit{Temporal Tensor Transformation Network} architecture. First, we introduce the key notations used in the paper as well as the formal problem definition for multivariate time series prediction. Then we introduce the \textit{Temporal Tensor Transformation} operation. Finally, we present the proposed neural network architecture. 

\subsection{Notations}
We use $x$ to denote a one-dimensional vector, $X$ to denote a two-dimensional matrix, and $\mathcal{X}$ to represent a three-dimensional tensor. Scalars with respect to their corresponding indices are denoted by lower-case letters, followed by sub-scripted letter triplets. For instance, $x_{i, j, k}$ of a 3D-tensor $\mathcal{X}$ indicates the \textit{(i, j, k)-th} scalar value of the tensor $\mathcal{X}$. Further notation and variables will be introduced and defined as they appear in the context.

\subsection{Problem Definition}
We now formally define the multivariate time series prediction problem. Given a complete multivariate time series dataset $X = \{ x_1, x_2, ..., x_n \}$, where $x \in \Bbb R^{m \times n}$, $m$ is the number of features, and $n$ is the total number of time steps in the multivariate series, our objective is to predict a future series of values up to a defined horizon window. Specifically, given a subset of the time series up to time step $T$, our model, denoted as function $F(\cdot)$, will take in an input series of $X_{T} = \{ x_1, x_2, ..., x_T\} \in \Bbb R^{m \times T}$, and subsequently generate an output sequence of $\hat{Y}_{T+h} \in \Bbb R^{m \times h}$, where $h$ is the horizon window. Hence, the model can be formulated as the mapping function $F(X_{T}) = \hat{Y}_{T+h}$.

\subsection{Temporal Tensor Transformation}
The proposed transformation augments the original data by adding a single higher-order dimension to the original data input. For example, if the input series data is a 1D vector of values (i.e., univariate time series), the transformed data structure will be a 2D matrix structure. Likewise, if the input is a 2D matrix (i.e., multivariate time series), the resulting structure will be a 3D tensor.

We define the Temporal Tensor Transformation as a mapping function $TT: X \rightarrow \tilde{\mathcal{X}}$, where $X \in \Bbb R^{m \times T}$ is the input multivariate time series and the resulting transformation generates a 3D tensor $\tilde{\mathcal{X}} \in \Bbb R^{m \times \omega \times o}$. Here, $\omega$ is the slice window size (i.e., the number of steps within a time window), and $o$ is the number of slices or the stack height of the resulting transformed tensor. The value of $o$ can be computed based on the hyperparameters. Our specific temporal tensor transformation is referred to as the \textit{Temporal-Slicing Stack Transformation}. Before introducing the Temporal-Slicing Stack operation, we will first introduce the hyperparameters involved in the slicing process. 


The \textit{window size}, denoted by $\omega$, defines the overall slicing window of the transformation function, and determines one of the major dimensions of the output tensor. The length of the slicing window is fixed by the number of features in the time series, i.e., $m$. Therefore, the resulting slicing window is of dimension $m \times \omega$. The \textit{stride} parameter of the slicing window, denoted by $s$, indicates how many time steps to advance the slicing window along the temporal dimension. The greater the stride, the lower the overall pattern resolution, due to the loss of information among contiguous values within the time series. The \textit{dilation} parameter $d$ is similar to the parameter introduced by Yu et al. \cite{yu2015multi} for dilated CNN, which allows us to slice a wider receptive window size without compromising to limited memory space.  The \textit{padding} value $p$ is similar to how CNN inherently increases the dimensionality of the original data structure. It allow for shift-invariant transformations as well as retaining the dimensional size of the data. Specifically, padding in our context  refers to the process of symmetrically appending a set of 1D-vectors of size $m$ to \textit{both} ends of the input time series matrix along the temporal dimension. However, we hypothesize that such values appended to the time series data may be potentially problematic, due to the risk of contaminating the original series data with noise or out-of-distribution values. Thus, we need to carefully choose the padding values, for example, using the same adjacent values or the local mean of vectors within a predetermined time window. 

Given the hyperparameters defined above, we can deterministically derive the number of slices $o$ as follows: \\
$$o = \Bigl\lfloor \dfrac{T + 2p - 2d(\omega - 1) - 1}{s} + 1\Bigr\rfloor$$

We summarize the Temporal-Slicing Stack Transformation in Algorithm \ref{alg}. For simplicity, we do not consider padding and dilation in this formulation (\textit{i.e. $p = 0$ and $d = 1$}).

\begin{algorithm}
\caption{Temporal-Slicing Stack Transformation}
\label{alg}
\begin{algorithmic}[1]
    \renewcommand{\algorithmicrequire}{\textbf{Input:}}
    \renewcommand{\algorithmicensure}{\textbf{Output:}}
    \REQUIRE $X \in \Bbb R^{m \times T}$: 2D input multivariate time series
    \ENSURE  $\mathcal{\tilde{X}} \in \Bbb R^{m \times \omega \times o}$: 3D output temporal tensor \\
 
    \textit{Init}: $\mathcal{\tilde{X}} \in \Bbb R^{m \times \omega \times o}$
    \FOR {$i = 1$ to $o$}
        \STATE $\mathcal{\tilde{X}}$[:, :, $i$] $\leftarrow X$[:, $i*s$:$i*s+\omega$] 
    \ENDFOR
    \RETURN $\mathcal{\tilde{X}}$
\end{algorithmic} 
\end{algorithm}

\begin{figure*}[ht!]
\centering
\includegraphics[width=0.975\textwidth]{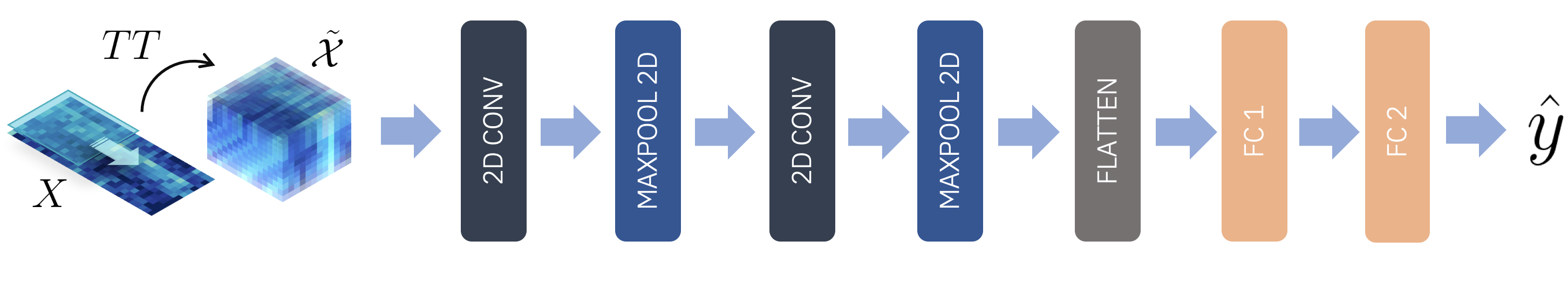}
\centering{\caption{Temporal Tensor Transformation Network Architecture}\label{architecture}}
\end{figure*}

\subsection{Neural Network Architecture}
In this section, we describe the deep learning architecture on top of the proposed temporal tensor transformation, which is referred to as \textit{TSSNet}. The neural network architecture is based on a fairly simple network structure, as shown in Figure \ref{architecture}. 
\subsubsection{Temporal-Slicing Stack Transformation}
Given the initial input multivariate time series $X$, we first perform the temporal tensor transformation as described in Algorithm \ref{alg}. As noted previously, the Temporal-Slicing Stack transformation is defined as a mapping of $TT: X \rightarrow \tilde{\mathcal{X}}$, where the input data is a two dimensional matrix $X \in \Bbb R^{m \times T}$ and the output is a three dimensional tensor $\tilde{\mathcal{X}} \in \Bbb R^{m \times \omega \times o}$.

\subsubsection{Convolutional Neural Network}
After transforming the input time series data to $\tilde{\mathcal{X}}$, we utilize a CNN to extract features from the tensor along the $\omega \times o$ plane. That is, we treat each feature as a separate channel. The total number of channels is therefore equal to the number of features $m$. We also fix one dimension of the CNN kernel to be the stack height $o$. The dimension of the kernel is thus $o \times k$, where $k$ is the width of the convolving kernel. Suppose we have $l$ CNN kernels, the $i$-th kernel learns the following set of weights:

$$h_{i} = W_{i} \ast \tilde{\mathcal{X}} + b_{i}, \hspace{0.5cm}i \in \{1,2,...,l\}$$

\noindent
where $\ast$ denotes the convolution operator, $W_{i}$ and $b_{i}$ are the weight and bias parameters, respectively. Note that during this process, we do not apply any sort of activation function as the model is sensitive to any type of operation that can collapse the range of values from the input (e.g., negative values will become zero in RELU). We empirically set the number of kernels to be $m$. The output feature maps from the convolution operation is subsequently fed into a max pooling operation, which is applied over the same $\omega \times o$ plane. We repeat these two operations twice in a successive manner.

\subsubsection{Dense Layer}
After the features are extracted from the CNN layer, the corresponding feature maps are then flattened out as a single vector. It is then subsequently fed into one fully connected hidden layer, $fc_1$:

$$fc_1 = W_1 \times h + b_1$$

\noindent
where $h$ is the 1D flatted out feature map from the previous CNN feature maps, and $W_1$ and $b_1$ are the weight and bias parameters, respectively. As a heuristic, the dimension of the hidden weight parameters used in this dense layer is usually greater than that of the output layer dimensions. Finally the subsequent latent vector $fc_1$ is fed into the output layer, where we learn the following set of parameters:

$$\hat{y} = W_2 \times fc_1 + b_2$$

\noindent
where $W_2$ and $b_2$ are the weight and bias parameters. The final output is a 1D vector, $\hat{y} \in \Bbb R^{m \times h}$, which can also be reshaped as a 2D matrix of dimensions $m \times h$.

\subsection{Objective Function}
To train the proposed neural network architecture, we minimize the squared error loss function:
$$\min_{\Theta} \; \sum_{t \in T_{train}} \| Y_t - \hat{Y_t}\|^{2}_{F}$$

\noindent
where the cost is minimized w.r.t. the parameters $\Theta$, and $\|\cdot\|^2_F$ denotes the Frobenius norm.

\subsection{Optimization Method}
To optimize the cost function, we utilize canonical methods for optimizing standard neural network. Specifically, we can apply common gradient-based methods such as stochastic gradient descent (SGD) or the Adam algorithm \cite{kingma2014adam}.
\section{Experiment}
We evaluate the performance of the proposed neural network architecture in this section. We first introduce several baseline models used for the benchmark comparison. We then introduce the evaluation metrics and data sets, and finally present the experimental results. 

\subsection{Baseline Models}
To compare the performance and robustness of our proposed model, we benchmarked against the following methods:

\begin{itemize}
    \item Vector Auto Regression (VAR)
    \item Long Short-Term Memory (LSTM)
    \item Gated Recurrent Unit (GRU)
    \item 1D Convolutional Neural Network (CNN)
    \item LSTNet \cite{lai2018modeling}
\end{itemize}

\noindent
All implementations were developed in PyTorch \cite{paszke2017automatic}, with the exception of the Vector Autoregression (VAR) model which is based on the Python StatsModels package \cite{seabold2010statsmodels}. our evaluations are performed across models which are stateful (i.e. VAR and RNN variants), stateless (i.e. CNN variants), and a hybrid of the two (i.e. LSTNet).

For Recurrent Neural Network variants, we implemented a many-to-one single layer vanilla architecture for the Long Short-Term Memory (LSTM) and Gated Recurrent Unit (GRU) with both having two fully connected layers appended at the end. Likewise, the 1D Convolutional Neural Network (CNN) architecture utilizes a single 1D CNN and a max pooling block with two fully connected layers. In this architecture, we treat the 2D input time series as a single channel image and perform convolution over the temporal-variable plane. Our LSTNet architecture is based on the code package from Lai et. al  \cite{lai2018modeling}. For all the neural network based models, they generate a batched multistep prediction values instead of a single time step output. For Vector Auto Regression, the predictions are generated based on an iterative manner, where the prediction at the current time step is appended to the previous input in order to predict the value at the next time step.

\subsection{Evaluation Metrics}
To compare all the methods, we use the following two evaluation metrics. The objective of our models is to minimize the Root Mean Squared Error (RMSE), while jointly maximizing the empirical correlation coefficient (CORR).\\

\noindent
\textbf{Root Mean Square Error (RMSE):} 
$$RMSE = \dfrac{1}{n} \sum_{i=1}^{n} \sqrt{\sum_{j=1}^{m} \sum_{t=1}^{h} (\mathcal{Y}_{ijt} - \hat{\mathcal{Y}}_{ijt})^2}$$

\noindent
where $\mathcal{Y}$, $\hat{\mathcal{Y}} \in \Bbb R^{n \times m \times h}$ such that $n$ is the total number of samples being computed. In this formulation we aggregate the error across all of the $m$ different feature for the total $h$ horizon of the predicted values.\\

\noindent
\textbf{Empirical Correlation Coefficient (CORR):}
$$CORR = \dfrac{1}{n} \sum_{i=1}^{n} 
    \dfrac{
        \sum_{j=1}^{m} \sum_{t=1}^{h} (\mathcal{Y}_{ijt} - \bar{\mathcal{Y}}_{i:t}) (\hat{\mathcal{Y}}_{ijt} - \bar{\hat{\mathcal{Y}}}_{i:t})
    }
    {
        \sqrt{\sum_{j=1}^{m} \sum_{t=1}^{h} (\mathcal{Y}_{ijt} - \bar{\mathcal{Y}}_{i:t})^2 (\hat{\mathcal{Y}}_{ijt} - \bar{\hat{\mathcal{Y}}}_{i:t})^2}
    }
$$

\noindent
where $\mathcal{Y}$, $\hat{\mathcal{Y}} \in \Bbb R^{n \times m \times h}$, $\bar{\mathcal{Y}}$, $\bar{\hat{\mathcal{Y}}}$, each denotes the ground truth value, the predicted value, the mean value of the ground truth for each feature, and the mean value of the predicted value for each feature, respectively. 

\subsection{Datasets}
\begin{figure}[ht]
\centering
\includegraphics[width=0.475\textwidth]{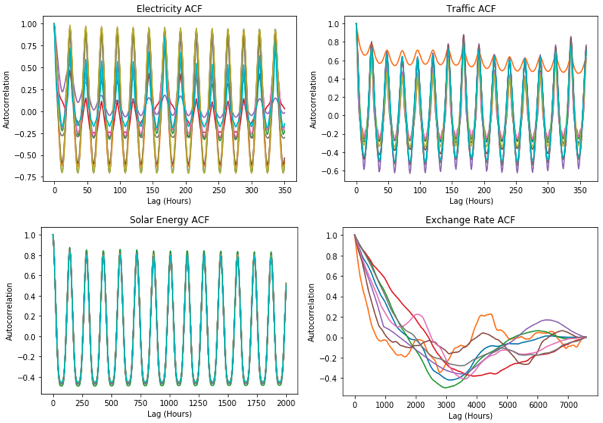}
\centering{\caption{Autocorrelation function plots of the raw datasets}\label{acf_plot}}
\end{figure}

We use four benchmark datasets across a wide variety of domain applications to compare the performance of all the models. Each dataset presented here has various degrees of non-stationarity. This allows us to evaluate the potential strengths and weaknesses of our proposed model architecture. Due to the memory-constraints, for each dataset, excluding the Exchange Rate dataset, we select at random ten features as a method of dimensionality reduction. We plot the autocorrelational functions for each dataset, as shown in Figure \ref{acf_plot}. 

\subsubsection{Electricity Usage \cite{UCI_electricity}}
The dataset contains electricity consumption recordings sampled at 15 min intervals between the year of 2012 to 2014 for 321 clients. The dataset was pre-processed to reflect the hourly consumption rate of the electricity usage instead. From the ACF plot, we see that different features have different levels of autocorrelation, notably a strong presence of seasonality as most of the features tend to have a consistent lag at around 24 hours.

\subsubsection{Foreign Exchange Rates}
The data set was collected over a period of 26 years between 1990 and 2016, and consists of the daily exchange rates from eight different countries. The dataset contains a total sample size of 7,588. Out of the four datasets, it does not show strong presence of the auto-correlational signal.

\subsubsection{Solar Power Production \cite{NREL_solar}}
The dataset comes from the Solar Power Data for Integration Studies, which contains the solar power and hourly day-ahead forecasts for approximately 6,000 simulated PV plants, at a sampling rate of every 5 minutes, in the year of 2006. From the ACF plot, we can see that the selected features all have uniform seasonality patterns with lag values around 140 hours.

\subsubsection{San Francisco Traffic \cite{PEMS_traffic}}
The dataset is from the Caltrans Performance Measurement System (PeMS), where traffic data was collected in real-time from over many detectors over the span of one year between 2015 and 2016, in the San Francisco Bay Area freeway. These detectors represent the density of the traffic between 0 and 1, where 0 indicates no traffic and 1 represents high amounts of congestion. From the ACF plot, we can see that most of the selected features have a seasonality with lag values around 160 hours.

For evaluation, we split the dataset into three different partitions in a chronological manner, with the first 60\% of the dataset will be used for training the model, the next 20\% of the data will be used for computing validation scores, and the remaining 20\% of the data will be used to test the model.

\subsection{Hyperparameters}
For training models, we tuned the hyperparameters using the \textit{Gaussian Process Optimization Framework (GPyOpt)} \cite{gpyopt2016}. We focused the parameter optimizations on the Temporal-Slicing Stack Transformation, in particular, the window size, $\omega$, between the range of 5 to 10, and the stride, $s$, between the range of 1 to 5. Furthermore, we also tuned the learning rate of our model between values of $[0, 0.01]$ and applied gradient clipping value of 10. Furthermore, we correspondingly adjusted the hidden weight dimensions of the model such that the dimension of the first dense layer had a greater number of parameters than the second output dense layer, or the final output dimension of the model. We perform 100 iterations of the modeling process and select the model with the highest empirical correlation coefficient score as our objective value to optimize over.

\subsection{Experimental Results}
\begin{table*}[ht!]
\caption{RMSE and CORR of Multivariate Time Series Prediction}
\label{results}

\centering
\resizebox{\textwidth}{!}{%

\begin{tabular}{|cc|cccc|cccc|cccc|cccc|}
\hline
\multicolumn{2}{|c|}{Dataset}        & \multicolumn{4}{c|}{Electricity}                                                                               & \multicolumn{4}{c|}{Solar Energy}                                                                              & \multicolumn{4}{c|}{Traffic}                                                                                   & \multicolumn{4}{c|}{Exchange Rate}                                                                             \\ \hline
Model                       & Metric & 15                        & 30                        & 60                        & 120                        & 15                        & 30                        & 60                        & 120                        & 15                        & 30                        & 60                        & 120                        & 15                        & 30                        & 60                        & 120                        \\ \hline
\multirow{2}{*}{VAR}        & RMSE   & 0.123                     & 0.135                     & 0.148                     & 0.162                      & 0.078                     & 0.129                     & 0.180                     & 0.219                      & 0.082                     & 0.089                     & 0.098                     & 0.108                      & \textbf{0.015}            & \textbf{0.021}            & \textbf{0.031}            & \textbf{0.047}             \\
                            & CORR   & 0.701                     & 0.667                     & 0.607                     & 0.507                      & 0.929                     & 0.809                     & 0.648                     & 0.425                      & 0.719                     & 0.679                     & 0.604                     & 0.477                      & \textbf{0.979}            & \textbf{0.961}            & \textbf{0.922}            & \textbf{0.845}             \\ \hline
\multirow{2}{*}{RNN (LSTM)} & RMSE   & 0.126                     & 0.125                     & 0.129                     & 0.131                      & 0.132                     & 0.190                     & 0.215                     & 0.219                      & 0.084                     & 0.087                     & 0.091                     & 0.094                      & 0.032                     & 0.034                     & 0.048                     & 0.057                      \\
                            & CORR   & 0.725                     & 0.709                     & 0.687                     & 0.672                      & 0.859                     & 0.643                     & 0.526                     & 0.517                      & 0.721                     & 0.694                     & 0.664                     & 0.646                      & 0.958                     & 0.942                     & 0.907                     & 0.837                      \\ \hline
\multirow{2}{*}{RNN (GRU)}  & RMSE   & \multicolumn{1}{l}{0.126} & \multicolumn{1}{l}{0.126} & \multicolumn{1}{l}{0.130} & \multicolumn{1}{l|}{0.131} & \multicolumn{1}{l}{0.132} & \multicolumn{1}{l}{0.192} & \multicolumn{1}{l}{0.214} & \multicolumn{1}{l|}{0.219} & \multicolumn{1}{l}{0.086} & \multicolumn{1}{l}{0.089} & \multicolumn{1}{l}{0.092} & \multicolumn{1}{l|}{0.094} & \multicolumn{1}{l}{0.021} & \multicolumn{1}{l}{0.030} & \multicolumn{1}{l}{0.037} & \multicolumn{1}{l|}{0.059} \\
                            & CORR   & \multicolumn{1}{l}{0.720} & \multicolumn{1}{l}{0.706} & \multicolumn{1}{l}{0.683} & \multicolumn{1}{l|}{0.668} & \multicolumn{1}{l}{0.857} & \multicolumn{1}{l}{0.634} & \multicolumn{1}{l}{0.522} & \multicolumn{1}{l|}{0.513} & \multicolumn{1}{l}{0.711} & \multicolumn{1}{l}{0.687} & \multicolumn{1}{l}{0.663} & \multicolumn{1}{l|}{0.640} & \multicolumn{1}{l}{0.971} & \multicolumn{1}{l}{0.945} & \multicolumn{1}{l}{0.911} & \multicolumn{1}{l|}{0.837} \\ \hline
\multirow{2}{*}{1D CNN}     & RMSE   & 0.057                     & 0.059                     & 0.073                     & 0.083                      & 0.085                     & 0.109                     & 0.123                     & 0.123                      & 0.070                     & 0.071                     & 0.069                     & 0.072                      & 0.039                     & 0.044                     & 0.051                     & 0.069                      \\
                            & CORR   & 0.891                     & 0.879                     & 0.831                     & 0.796                      & 0.946                     & 0.918                     & 0.899                     & 0.885                      & 0.837                     & 0.831                     & 0.835                     & 0.835                      & 0.937                     & 0.909                     & 0.847                     & 0.729                      \\ \hline
\multirow{2}{*}{LSTNet}     & RMSE   & 0.064                     & 0.062                     & 0.060                     & 0.060                      & 0.077                     & 0.094                     & 0.110                     & 0.119                      & 0.066                     & 0.067                     & 0.068                     & 0.068                      & 0.056                     & 0.059                     & 0.059                     & 0.074                      \\
                            & CORR   & 0.890                     & 0.867                     & 0.873                     & 0.868                      & 0.956                     & 0.936                     & 0.906                     & 0.899                      & 0.855                     & 0.847                     & 0.843                     & 0.843                      & 0.815                     & 0.796                     & 0.747                     & 0.699                      \\ \hline
\multirow{2}{*}{TSSNet}     & RMSE   & \textbf{0.055}            & \textbf{0.057}            & \textbf{0.059}            & \textbf{0.058}             & \textbf{0.076}            & \textbf{0.094}            & \textbf{0.105}            & \textbf{0.098}             & \textbf{0.062}            & \textbf{0.062}            & \textbf{0.067}            & \textbf{0.065}             & 0.029                     & 0.040                     & 0.050                     & 0.065                      \\
                            & CORR   & \textbf{0.896}            & \textbf{0.888}            & \textbf{0.879}            & \textbf{0.878}             & \textbf{0.956}            & \textbf{0.937}            & \textbf{0.918}            & \textbf{0.909}             & \textbf{0.859}            & \textbf{0.860}            & \textbf{0.852}            & \textbf{0.855}             & 0.944                     & 0.914                     & 0.895                     & 0.780                      \\ \hline
\end{tabular}%

}

\end{table*}
\begin{figure*}[ht]
    \includegraphics[width=\textwidth]{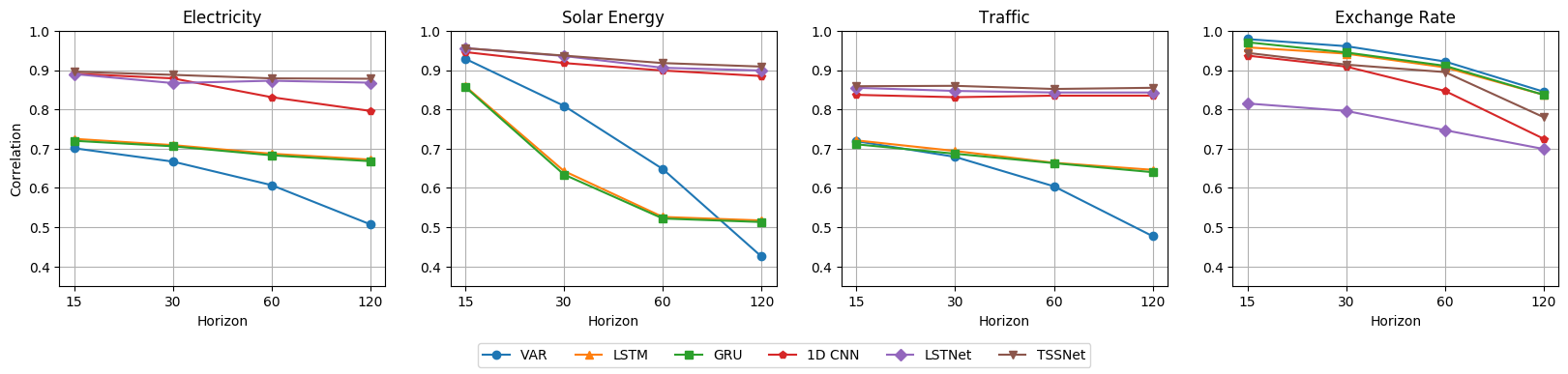}
    \caption{Empirical correlation coefficient plots by  prediction horizon}
    \label{corr_plot}
\end{figure*}

\begin{figure}[ht]
\centering
\includegraphics[width=0.48\textwidth]{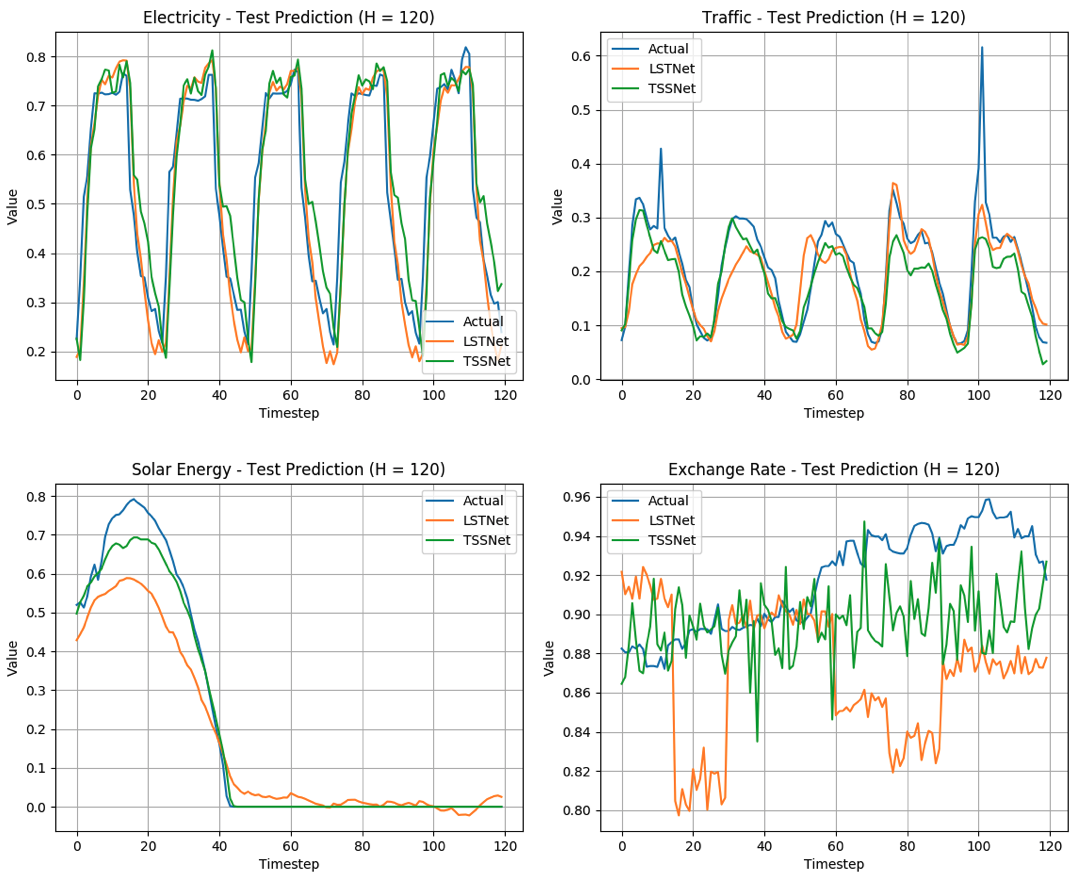}
\centering{\caption{Comparison of TSSNet and LSTNet predictions}\label{pred_plot}}
\end{figure}

We present the experimental results on the two defined evaluation metrics in Table \ref{results}. For each dataset, we provide the model a fixed input of $T=168$ time steps and obtained predictions against a horizon window of 15, 30, 60, and 120 time steps. We note that these horizons are multi-time step windowed predictions, as opposed to a single-point estimation. The best performing models with respect to the empirical correlations are in bold for each dataset and horizon. We also provide a comparative graph summary of the Empirical Correlation Coefficient as shown in Figure \ref{corr_plot}.

Based on the empirical results, our proposed model (denoted by TSSNet) outperforms the baseline metrics for the Electricity, Solar Energy, and Traffic datasets, and outperforms some of the models for the Exchange Rates dataset. We attribute our analysis based on two key observations: i) the use of stateful versus stateless model types and ii) the autocorrelations present in the dataset.

In Figure \ref{corr_plot}, we observe a noticeable cluster between stateless models (i.e. 1D CNN, LSTNet, and TSSNet) and stateful models (i.e. LSTM, GRU, and VAR) in terms of the overall performance. Furthermore, we observe a sharper performance degradation over larger temporal horizon for stateful models than stateless models. Within stateless models, our proposed model consistently retains an upper bound performance, while the 1D CNN model primarily remains as a lower bound with the LSTNet in between. This is also present in the Exchange Rate dataset.

The other factor affecting the overall performance of our model is the presence of autocorrelations in the dataset, in particular, the seasonality. As previously shown in Figure \ref{acf_plot}, we have noted that there is a strong presence of autocorrelationa within the Electricity, Solar Energy, and Traffic datasets, while weaker in the Exchange Rate dataset. This realization in context with our empirical results shed light on some observations. For highly autocorrelational data, stateless models tend to perform better over stateful models. However, with weaker autocorrelational signals such as the Exchange Rate dataset, our model performs better than the stateless models, but marginally worse than the stateful model variants.

In particular, our experimental results with 1D CNNs and LSTNet demonstrate that the methods in how convolution is applied over the temporal data can lead to difference in performance. Hence, this comparison indicates that our transformation method enables high efficiency of information gain with respect to the autocorrelational features captured over a large temporal region.

We also plot sample predictions against our test set at horizons $h=120$, with an empirical comparison against LSTNet. The results are shown in Figure \ref{pred_plot}. We have chosen to evaluate LSTNet for its marginally close performance in terms of the empirical correlation coefficient, but having slightly lower RMSE scores as shown in Table \ref{results}. From Figure \ref{pred_plot}, we can observe how effectively our model can predict non-linear patterns as the model is able to adapt to most of the acute signals in the data. From the Electricity dataset, we observe much more precise adaptations to the peaks and troughs of the cyclical signals. In the Traffic dataset, the proposed TSSNet is able to capture the sinusoidal patterns, however is slightly less precise to sudden peaks such as those found in the last two cycles of the dataset. In the Solar Energy dataset, we observe a closer fit to the overall shape of the data, as it is able to very closely capture both the non-linear and linear components of the data. However, for the Exchange Rate dataset, we see that both models have a relatively hard time with the predictions due to the high noisy outputs. In TSSNet, we observe that the predictions are centralized around a consistent range of values, while LSTNet's predictions are sporadically shifting the prediction values from time to time.
\section{Sensitivity Analysis}
\begin{figure*}[ht]
    \centering
    \includegraphics[width=0.8\textwidth]{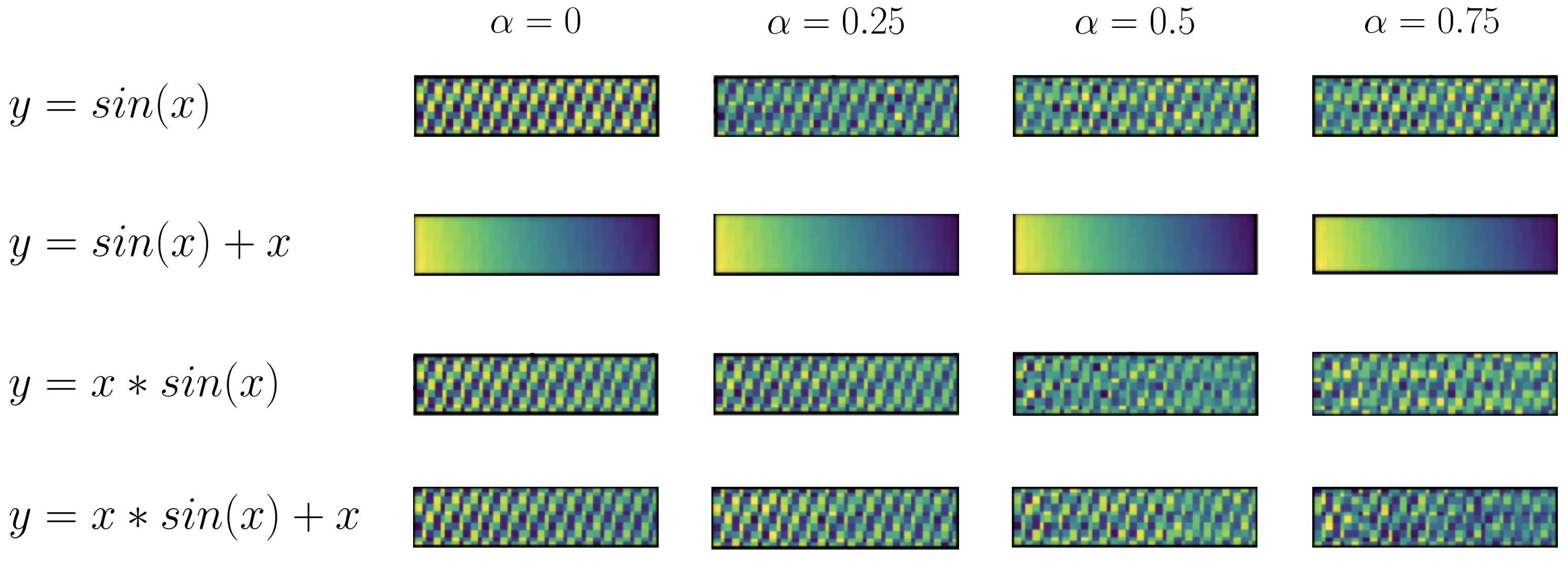}
    \caption{CNN feature maps of synthesized function datasets} \label{featmap}
\end{figure*}

In this section, we perform sensitivity analysis to evaluate and understand the advantages and limitations of our proposed architecture. This analysis fixates the intermediate model architecture and manipulates the input and output data parameters in a controlled manner. We observe the resulting effects to the overall performance of the model and draw conclusions on the robustness of our architecture.

In particular, our objectives in this analysis are two folds. First, we empirically study and demonstrate the underlying properties of how our proposed model is able to learn emergent autocorrelational patterns from controlled and synthesized inputs. We subsequently analyze the activation maps to better understand the extracted features. In our second analysis, we evaluate the robustness of the proposed model architecture by varying the input and output dimensions. These two experiments will help us better identify some of the core inner workings and draw insights of the model.

\subsection{Synthesized Data Sensitivity Analysis}

In this section, we describe methods used for observing the underlying feature maps that are generated by CNN. We perform these experiments to empirically identify properties of autocorrelational signals such as seasonality and trend. For this experiment, we generated a set of synthetic datasets from simple functions which explicitly exhibit properties of seasonality and trend, as demonstrated in Figure \ref{featmap}. Through these controlled and simplified experiments, we can uncover some of the underlying mechanisms for extracting autocorrelational features from the data.

We overfit our models with inputs from known simple deterministic functions, such as the sine and linear functions. We then evaluate the resulting activation maps (after the first convolutional layer) generated by the model, in order to identify which salient features are extracted by the learned CNN filters. Furthermore, to evaluate robustness of our model, we inject noise of various degrees by adding a noise factor $\epsilon$ within the sine function argument as $sin(x + \alpha \epsilon)$, where $\alpha$ is the ratio of the noise present, and $\epsilon \in \Bbb R$ is a random number sampled from a standard normal distribution. For each function, we empirically evaluate the robustness by increasing the degree of $\alpha$ from 0 to 0.75.

Based on the experimental formulation above, we demonstrate the feature maps generated by the CNN layer, as presented in Figure \ref{featmap}. Along the vertical axis, we list the different deterministic functions utilized to generate the synthetic dataset, while on the horizontal axis we show the effect over the different rates of noise injection.

Empirically, we found that the CNN filters perform the roles of both a feature extractor as well as a de-noising component of the model. For the first function, $y = sin(x)$, we can observe that the model is able to pick up the uniform repetitive pattern of the data as exhibited by the uniform checkerboard like pattern. Even with the injected noise, we see that the filter is able to de-noise a fair amount of the artifacts from the model and still retain the core checkerboard-like pattern as exhibited when $\alpha = 0$. In the second function, $y = sin(x) + x$, we observe across all values of $\alpha$, the activation map shows a uniform gradient-like pattern. This indicates that the model attenuates the $sin(x)$ component of the function and focuses only on the linear component of the model. In essence, the filter treats $sin(x)$ as a constant and emphasizes the monotonically increasing nature of the time series data from the linear $x$ component of the equation. Hence, the resulting activation map reveals a gradient like pattern which indicates an increasing trend pattern. For the third and fourth functions, we observe the same uniform checkerboard pattern similar to the first function. Furthermore, we observe a slight gradient effect indicating that the values on the left region of the plot is darker and the right region is lighter. This implies that the model's learned filter can extract both the seasonal component as well as the linear trend pattern concurrently from the synthetic dataset. The fourth function contains a slightly darker pattern which is indicative of the extra linear factor.

From these controlled experiments, we empirically demonstrate the effectiveness of using CNNs as feature extractors to identify core autocorrelational components of the model, notably, seasonality and trend from the transformed temporal tensor representations. Furthermore, we also demonstrate that such learned filters are able to act as de-noising filters, which can enable generalizations for noisy input time series.

\subsection{Sensitivity Analysis on Input vs Prediction Horizon}
In our previous experiments, we have only evaluated our models against a fixed input size of 168 time steps  while we correspondingly varied the output horizon time steps for prediction. In this experiment, we perform a sensitivity analysis by varying both the input and output dimensions. One key motivation behind this experimentation is to assess the balance of the input and output size and how they influence the overall predictive power of the model. Various degrees of temporal granularity, such as daily, weekly, or monthly perspectives of the input data can strongly influence the respective outcome of the prediction due to the observational scope of the information provided to the model. This is particularly important when considering autocorrelational patterns, as long-range dependencies of repetitive patterns can significantly change with respect to the different temporal granularities of the data. As a result, the representation the model learns will also differ and consequently affect the prediction performance. For this analysis, we constrain our experiment to only evaluate the performance of our proposed model architecture (i.e. TSSNet). We vary the input sizes by 32, 64, 128, and 256, while also vary the output horizon sizes by 15, 30, 60, and 120, respectively. We perform experiments with every combination of these input and output dimensions for evaluation.

\begin{table*}[ht!]
\caption{Input vs Horizon Sensitivity Analysis Results}
\label{results_2}

\centering
\resizebox{\textwidth}{!}{

\begin{tabular}{cccccccccccccccccc}
\hline
\multicolumn{2}{|c|}{Dataset}                                             & \multicolumn{4}{c|}{Electricity}                                                                               & \multicolumn{4}{c|}{Solar Energy}                                                                              & \multicolumn{4}{c|}{Traffic}                                                                                   & \multicolumn{4}{c|}{Exchange Rate}                                                                             \\ \hline
\multicolumn{1}{|c|}{Input Size}           & \multicolumn{1}{c|}{Horizon} & 15                        & 30                        & 60                        & \multicolumn{1}{c|}{120}   & 15                        & 30                        & 60                        & \multicolumn{1}{c|}{120}   & 15                        & 30                        & 60                        & \multicolumn{1}{c|}{120}   & 15                        & 30                        & 60                        & \multicolumn{1}{c|}{120}   \\ \hline
\multicolumn{1}{|c|}{\multirow{2}{*}{32}}  & \multicolumn{1}{c|}{RMSE}    & 0.091                     & 0.098                     & 0.105                     & \multicolumn{1}{c|}{0.109} & 0.091                     & 0.128                     & 0.145                     & \multicolumn{1}{c|}{0.144} & 0.068                     & 0.072                     & 0.077                     & \multicolumn{1}{c|}{0.075} & 0.018                     & 0.027                     & 0.036                     & \multicolumn{1}{c|}{0.038} \\
\multicolumn{1}{|c|}{}                     & \multicolumn{1}{c|}{CORR}    & 0.862                     & 0.835                     & 0.810                     & \multicolumn{1}{c|}{0.796} & 0.941                     & 0.879                     & 0.837                     & \multicolumn{1}{c|}{0.840} & 0.846                     & 0.825                     & 0.800                     & \multicolumn{1}{c|}{0.804} & 0.974                     & 0.949                     & 0.919                     & \multicolumn{1}{c|}{0.918} \\ \hline
\multicolumn{1}{|c|}{\multirow{2}{*}{64}}  & \multicolumn{1}{c|}{RMSE}    & 0.094                     & 0.100                     & 0.104                     & \multicolumn{1}{c|}{0.110} & 0.099                     & 0.106                     & 0.113                     & \multicolumn{1}{c|}{0.115} & 0.067                     & 0.071                     & 0.072                     & \multicolumn{1}{c|}{0.071} & 0.024                     & 0.029                     & 0.041                     & \multicolumn{1}{c|}{0.057} \\
\multicolumn{1}{|c|}{}                     & \multicolumn{1}{c|}{CORR}    & 0.850                     & 0.828                     & 0.811                     & \multicolumn{1}{c|}{0.809} & 0.927                     & 0.917                     & 0.905                     & \multicolumn{1}{c|}{0.907} & 0.848                     & 0.829                     & 0.818                     & \multicolumn{1}{c|}{0.823} & 0.960                     & 0.942                     & 0.899                     & \multicolumn{1}{c|}{0.824} \\ \hline
\multicolumn{1}{|c|}{\multirow{2}{*}{128}} & \multicolumn{1}{c|}{RMSE}    & \multicolumn{1}{l}{0.091} & \multicolumn{1}{l}{0.101} & \multicolumn{1}{l}{0.103} & \multicolumn{1}{l|}{0.101} & \multicolumn{1}{l}{0.083} & \multicolumn{1}{l}{0.093} & \multicolumn{1}{l}{0.101} & \multicolumn{1}{l|}{0.102} & \multicolumn{1}{l}{0.066} & \multicolumn{1}{l}{0.068} & \multicolumn{1}{l}{0.071} & \multicolumn{1}{l|}{0.069} & \multicolumn{1}{l}{0.029} & \multicolumn{1}{l}{0.032} & \multicolumn{1}{l}{0.034} & \multicolumn{1}{l|}{0.063} \\
\multicolumn{1}{|c|}{}                     & \multicolumn{1}{c|}{CORR}    & \multicolumn{1}{l}{0.861} & \multicolumn{1}{l}{0.828} & \multicolumn{1}{l}{0.818} & \multicolumn{1}{l|}{0.829} & \multicolumn{1}{l}{0.950} & \multicolumn{1}{l}{0.937} & \multicolumn{1}{l}{0.926} & \multicolumn{1}{l|}{0.923} & \multicolumn{1}{l}{0.845} & \multicolumn{1}{l}{0.836} & \multicolumn{1}{l}{0.832} & \multicolumn{1}{l|}{0.835} & \multicolumn{1}{l}{0.954} & \multicolumn{1}{l}{0.936} & \multicolumn{1}{l}{0.887} & \multicolumn{1}{l|}{0.802} \\ \hline
\multicolumn{1}{|c|}{\multirow{2}{*}{256}} & \multicolumn{1}{c|}{RMSE}    & 0.093                     & 0.097                     & 0.101                     & \multicolumn{1}{c|}{0.101} & 0.075                     & 0.089                     & 0.101                     & \multicolumn{1}{c|}{0.106} & 0.063                     & 0.064                     & 0.066                     & \multicolumn{1}{c|}{0.068} & 0.046                     & 0.054                     & 0.063                     & \multicolumn{1}{c|}{0.073} \\
\multicolumn{1}{|c|}{}                     & \multicolumn{1}{c|}{CORR}    & 0.857                     & 0.854                     & 0.821                     & \multicolumn{1}{c|}{0.826} & 0.958                     & 0.942                     & 0.925                     & \multicolumn{1}{c|}{0.919} & 0.856                     & 0.854                     & 0.844                     & \multicolumn{1}{c|}{0.832} & 0.886                     & 0.858                     & 0.773                     & \multicolumn{1}{c|}{0.715} \\ \hline
\multirow{2}{*}{}                          &                              &                           &                           &                           &                            &                           &                           &                           &                            &                           &                           &                           &                            &                           &                           &                           &                            \\
                                           &                              &                           &                           &                           &                            &                           &                           &                           &                            &                           &                           &                           &                            &                           &                           &                           &                            \\
\multirow{2}{*}{}                          &                              & \textbf{}                 & \textbf{}                 & \textbf{}                 & \textbf{}                  & \textbf{}                 & \textbf{}                 & \textbf{}                 & \textbf{}                  & \textbf{}                 & \textbf{}                 & \textbf{}                 & \textbf{}                  &                           &                           &                           &                            \\
                                           &                              & \textbf{}                 & \textbf{}                 & \textbf{}                 & \textbf{}                  & \textbf{}                 & \textbf{}                 & \textbf{}                 & \textbf{}                  & \textbf{}                 & \textbf{}                 & \textbf{}                 & \textbf{}                  &                           &                           &                           &                           
\end{tabular}%

}

\end{table*}
\begin{figure*}[ht]
    \includegraphics[width=\textwidth]{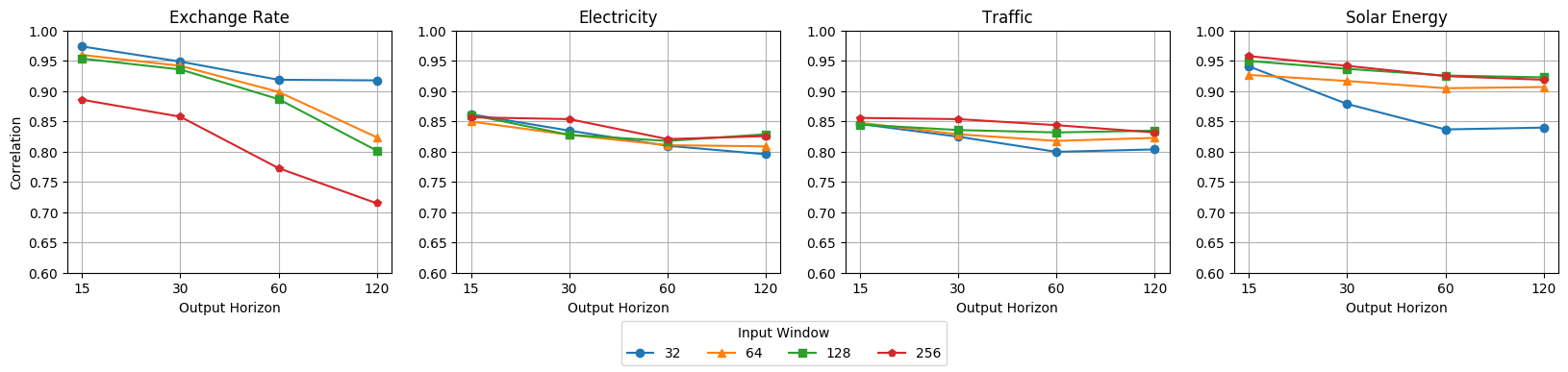}
    \caption{Input size vs output horizon correlation degradation}
    \label{results_2a}
\end{figure*}

\begin{figure*}[ht]
    \includegraphics[width=\textwidth]{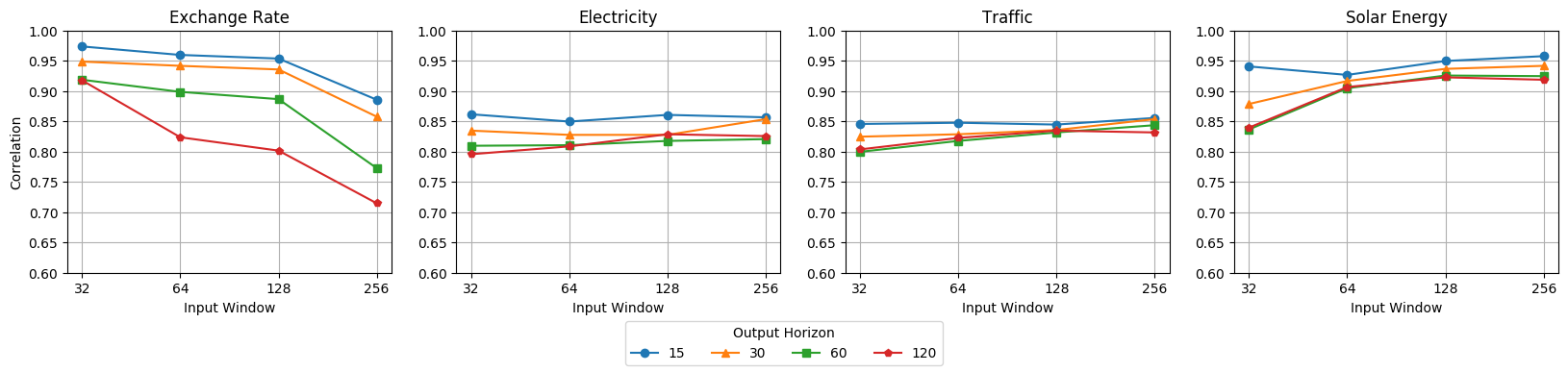}
    \caption{Output horizon vs input size correlation degradation}
    \label{results_2b}
\end{figure*}

The experimental results are shown in Table \ref{results_2}. We also provide two plots, which evaluate the empirical correlation coefficient performance degradation from the perspective of the input and output dimensions respectively, as shown in Figures \ref{results_2a} and \ref{results_2b}. From these plots, we can draw upon several key insights regarding the overall robustness of our model, as well as some heuristics which can potentially help with improving the overall performance of the model.

In Figure \ref{results_2a}, we present a plot which demonstrates how the empirical correlation coefficient performance with different input dimensions degrades with respect to the different output horizon dimensions. First, we can observe a similar performance degradation effect shown from our previous experiments. In particular, one key observation to note is the variance of the correlation with respect to each of the different output horizons are greater when the data is not highly auto-correlated. For example, the Exchange Rate time series  does not have strong autocorrelation, therefore exhibiting a wider variance of correlation scores for the input window size with respect to the output horizon dimensions. In contrast, for datasets such as the Electricity and Traffic, the overall correlation performance degradation is relatively smaller and its variance bounds are significantly tighter. However, one detail to make note of in Figure \ref{results_2a} is the top performing models for each of the different datasets. While in the Exchange Rate dataset, the model with input window of 32 performs the best, the other datasets mostly have larger input window sizes performing better in general. This results suggest that the overall input size can significantly affect the performance of the model. This also provides further evidence that our proposed model can better extract features from data which exhibits long-range autocorrelational patterns.

In Figure \ref{results_2b}, we present a different view of the results which demonstrates how the empirical correlation coefficient performance with different output dimensions is influenced by varying the input sizes. The motivation behind this analysis allows us to evaluate the overall robustness of the different output horizon dimensions of the model's prediction with respect to varying the overall input window dimensions. Unlike the previous analysis from our first experiment, we focus on empirically evaluating the effect of the benefited information gain the model has obtained when we provide a greater amount of temporal information (i.e., a larger input window). In Figure \ref{results_2b}, we also notice very similar effects of the variance bounds of the correlation results with respect to the different input window sizes, as highly autocorrelated data tend to have smaller bounds while non-autocorrelated data have larger variance ranges. In particular, increasing the input window sizes with respect to the different output horizon dimensions, we note that the overall performance degradation is minimal, and in some cases performance improves for datasets which exhibit high autocorrelational properties. In contrast, when the data does not contain any sort of autocorrelational signals (e.g., the Exchange Rate), the model performance drops significantly as we increase the input window sizes. These results empirically reinforce the notion that our model is able to robustly learn long-range temporal dependencies with high autocorrelational features.
\section{Conclusion}
We propose a neural network architecture for multivariate time series prediction through a new class of transformation function known as Temporal Tensor Transformations. In particular, we present the Temporal-Slicing Stack Transformation which utilizes a slice-based operator to transform the original 2D time series into a 3D tensor. This transformation encodes long-range auto-correlational features that can be extracted by a Convolutional Neural Network. Both experimental results and sensitivity analysis provide strong evidence that the proposed architecture is able to learn non-linear auto-correlational patterns effectively from the data.

For future work, we plan to investigate various components of the proposed architecture. To further understand the underlying mechanisms and sensitivity of the architecture, we aim to carefully identify which hyperparameters are more sensitive with respect to the overall model performance. Additionally, to better improve the prediction performance of the model on data without strong autocorrelation, we will explore the use of hybrid approaches that combine both our stateless approach in feature extraction in conjunction with recurrent neural networks. As an extension to the proposed temporal tensor transformation, we can also explore other types of transformation methods which utilize explicit components, such as multivariate interactions and variable sampling rates among different variables. The core challenges behind these classes of transformation functions lie in how to design new structures that can enable efficient information gain through the use of CNN for feature extractions.

\bibliographystyle{IEEEtran}
\bibliography{reference.bib}

\end{document}